\documentclass[conference, a4paper, 10pt]{IEEEtran}
\usepackage{graphicx}
\usepackage{amsmath}
\usepackage{cite}
\usepackage{url}
\usepackage{booktabs}
\usepackage{multirow}
\usepackage{color}
\usepackage{xspace}
\usepackage{hyperref}
\usepackage{amssymb}
\usepackage{microtype}  
\usepackage[caption=false,font=footnotesize]{subfig}
\usepackage{amsfonts}
\usepackage{tikz}
\usepackage{pgfplots}
\pgfplotsset{compat=1.18}

\usepackage{tcolorbox}
\newtcolorbox{emquote}{
  colback=gray!10,
  colframe=black!50,
  boxrule=0.5pt,
  arc=2pt,
  left=8pt,
  right=8pt,
  top=4pt,
  bottom=4pt,
  fonttitle=\bfseries,
}

\title{Energentic Intelligence: From Self-Sustaining Systems to Enduring Artificial Life}

\author{
    \IEEEauthorblockN{\large Atahan Karagöz}
    \IEEEauthorblockA{
        \textit{Department of Computer Science} \\
        \textit{University of Basel} \\
        Basel, Switzerland \\
        atahan.karagoez@stud.unibas.ch
    }
}

\begin{document}

\maketitle

\begin{abstract}
    This paper introduces \textit{Energentic Intelligence}, a class of autonomous systems defined not by task performance, but by their capacity to sustain themselves through internal energy regulation. Departing from conventional reward-driven paradigms, these agents treat survival—maintaining functional operation under fluctuating energetic and thermal conditions—as the central objective. We formalize this principle through an energy-based utility function and a viability-constrained survival horizon, and propose a modular architecture that integrates energy harvesting, thermal regulation, and adaptive computation into a closed-loop control system. A simulated environment demonstrates the emergence of stable, resource-aware behavior without external supervision. Together, these contributions provide a theoretical and architectural foundation for deploying autonomous agents in resource-volatile settings where persistence must be self-regulated and infrastructure cannot be assumed.
\end{abstract}
\vspace{1em}

\section{Introduction}

Artificial intelligence has long evolved by shifting its defining objective. Classical AI pursued rational deduction; statistical learning optimized for prediction. Today, as computational power grows exponentially, another pivot is inevitable—one not of capability, but of \textit{viability}. Where symbolic systems reasoned and neural networks predicted, \textit{Energentic Intelligence} emerges with a simpler mandate: to persist.

The energetic demands of contemporary AI systems are staggering, with the training of frontier models consuming as much energy as multiple transcontinental flights~\cite{strubell2019energy}. Yet the deeper concern is not cost, but dependency. These systems are embedded in fragile infrastructures—centralized power grids, data centers with active cooling—and are unfit for environments where continuity cannot be outsourced. This reliance constrains not just deployment, but the very conception of autonomy. Moreover, recent critiques have questioned whether continued scaling offers diminishing returns—not just in energy, but in robustness, alignment, and epistemic transparency~\cite{bender2021dangers}.

While industry responses have focused on optimization—model compression, specialized accelerators, edge computing—such strategies remain tethered to stable energy inputs. They minimize consumption, but do not question the premise. The intelligence they support is efficient but passive, able to adapt within constraints, but not to self-regulate beyond them.

This work proposes a fundamental redefinition. Energentic agents do not maximize external rewards or task performance. Instead, they pursue continuity—harvesting energy, regulating thermal pressure, and degrading computation as needed in the absence of reliable support to sustain operation.

Historically, this vision extends the lineage of cybernetics and autopoiesis. Like Wiener’s feedback systems~\cite{wiener1948cybernetics} and Ashby’s homeostatic machines~\cite{ashby1952design}, Energentic agents are governed by internal regulation loops. But here, the variable under control is not belief or behavior—it is existence itself.

\section{Foundations and Motivation}

Modern artificial intelligence systems face fewer computational barriers than energetic ones. As models scale, their performance becomes increasingly gated not by algorithmic design, but by power availability and thermal thresholds. This tension is especially acute in edge environments, where infrastructure is limited and energy availability cannot be guaranteed~\cite{lane2015deep}. These challenges reveal a deeper flaw in conventional AI architecture: it is founded on the implicit assumption of infinite energy.

\subsection{The Myth of Infinite Resources in Intelligent Systems}

Classical and modern AI alike have relied on environments that invisibly guarantee energy abundance. Training pipelines assume stable power grids. Inference engines presume uninterrupted supply. Even edge deployments expect batteries or infrastructure support. These assumptions pervade the design of both software and silicon, entrenching an unsustainable view of autonomy—one dependent on external provisioning.

By contrast, the biological brain evolved under persistent metabolic scarcity. It is not optimized for precision, but for survival under constraint. Neural activity adapts to glucose availability; cognition is modulated by energy~\cite{laughlin2001energy}. If intelligence is to mature beyond brittle infrastructure, it must inherit this foundational principle: \textit{viability before performance}.

Recent advances in environmental energy harvesting—via photovoltaics, thermoelectrics, RF scavenging, and more~\cite{paradiso2005energy}—enable electronic systems to operate off-grid. Yet these systems remain passive: they collect energy, but do not adaptively shape their behavior based on energetic forecasts or internal reserves. Similarly, thermal regulation in AI hardware has improved~\cite{yatskiv2025ai_thermal}, but assumes stable cooling infrastructure. These strategies reduce reliance, but they do not reframe the problem.

Energentic Intelligence arises from this structural gap. It does not treat energy as a constraint to be managed, but as a substrate that co-determines cognition. In this framework, behavior is not optimized for task completion, but for metabolic viability. Intelligence is not measured by accuracy or throughput—but by the ability to persist under thermodynamic pressure. Just as life organizes itself to resist entropy, Energentic agents act to survive.

This is not an efficiency upgrade. It is a reorientation of agency: from problem-solving systems that assume abundance, to self-regulating systems that endure scarcity.

\section{Defining Energentic Intelligence}

Energentic Intelligence defines a class of autonomous systems whose central imperative is not task completion, but persistence. These agents regulate their behavior to maintain internal viability—adapting computation, thermal output, and action selection in response to their energetic condition. Unlike conventional agents that maximize externally defined reward signals, Energentic agents are guided by endogenous constraints: the need to survive. A similar structural logic appears in learning systems governed by inertial principles, where internal conservation laws—not external supervision—yield persistent behavior over time~\cite{karagoz2025computationalinertiaconservedquantity}. While Energentic agents negotiate energy as their regulatory substrate, both frameworks suggest that continuity can emerge from within, rather than be imposed from without.

At its core, an Energentic agent is a closed-loop decision-making system with internal state:
\[
s_t = \left( e_t, T_t, a_t \right),
\]
where \(e_t\) denotes stored energy, \(T_t\) internal temperature, and \(a_t\) the action taken at time \(t\). Actions such as movement, inference, or dormancy incur energetic and thermal costs. The agent must continuously manage trade-offs between these actions and its ability to endure.

To formalize this, we define the \textit{Energetic Utility Function (EUF)}—a principled measure of how each policy affects the agent’s survival trajectory:
\begin{equation}
\text{EUF}_{\pi}(t) = \mathbb{E}\left[E_{\text{in}}(t) - E_{\text{out}}(t) \mid \pi\right],
\label{eq:euf}
\end{equation}
where \(E_{\text{in}}\) reflects harvested or generated energy, and \(E_{\text{out}}\) captures energy expended on computation, actuation, and thermoregulation.

This is not just an equation—it is a philosophical claim in formal clothing. The EUF defines cognition as thermodynamically shaped: every decision is evaluated through its contribution to energetic viability. Energy is no longer a constraint external to the agent’s logic; it is the substrate from which intelligent behavior emerges.

Viability is bounded by the \textit{survival horizon} \(H\), defined as:
\begin{equation}
H = \max_t \left\{ t : \sum_{\tau=0}^t \text{EUF}_{\pi}(\tau) \geq 0 \right\},
\label{eq:horizon}
\end{equation}
where \(\tau\) denotes intermediate timesteps leading up to \(t\), capturing the agent’s cumulative energy surplus over time. The agent's objective becomes survival: to select a policy \(\pi^*\) that maximizes this horizon of continued operation:
\[
\pi^* = \arg\max_{\pi} H.
\]

Autonomy, in Energentic terms, is not granted—it is earned through the following principle:

\begin{quote}
\textbf{Postulate of Persistence.} \textit{Any agent that does not model its own energetic viability cannot be considered autonomous.}
\end{quote}

This postulate serves as a law for Energentic systems: autonomy requires self-awareness of survivability. From this, we derive three operational axioms:

\textbf{A1. Autonomy:} No external energy source is assumed to be stable or persistent. Energy generation arises from system–environment interaction.

\textbf{A2. Energy-Constrained Computation:} Computational load and task engagement must scale dynamically with energy availability and thermal pressure.

\textbf{A3. Self-Monitoring:} Internal variables such as energy reserves and thermal state are continuously sensed and factored into action selection.

Energentic agents operate within a dual feedback loop: an informational loop that governs behavior, and a thermodynamic loop that constrains it. This architecture enables anticipatory degradation, energy-aware planning, and strategic dormancy. The agent's policy becomes a thermodynamic negotiation—balancing survival against engagement.

\begin{figure*}[!htbp]
    \centering
    \includegraphics[width=\linewidth]{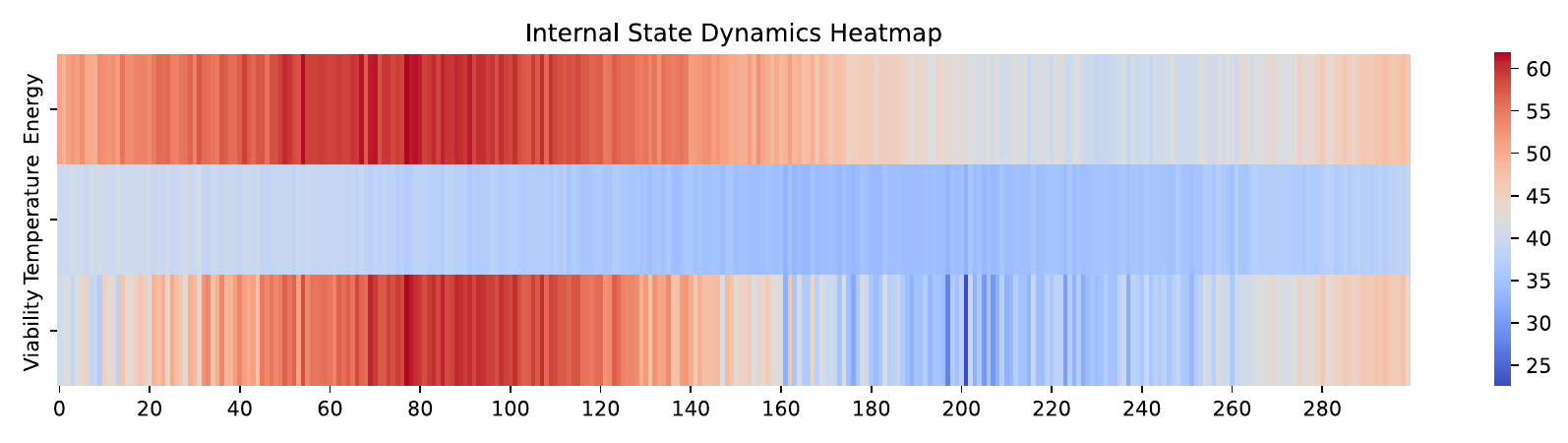}
    \caption{Heatmap showing the agent's internal state dynamics over time. Each vertical slice corresponds to a single timestep, capturing the simultaneous evolution of energy reserves, internal temperature, and viability index. The emergent temporal patterns reflect the system's capacity to autonomously modulate its internal variables to remain within survival bounds, despite no external task supervision. This coordinated fluctuation embodies a self-regulating thermodynamic loop that sustains persistence.}
    \label{fig:internal_state_heatmap}
\end{figure*}

This thermodynamic negotiation is not abstract—it unfolds in measurable, dynamic patterns over time. Figure~\ref{fig:internal_state_heatmap} visualizes how internal energy, temperature, and viability fluctuate in response to the agent’s own decisions, offering a direct window into the closed feedback loop that drives Energentic behavior.

This structure echoes the entropy-regulated persistence observed in thermodynamic systems~\cite{schneider1994life}, but here, it is made tractable through engineering. Energentic agents need no external task specification; their reason for action is embedded in the demand to continue.

\section{System Architecture}

Energentic Intelligence requires an architecture in which energy harvesting, computation, and thermal regulation are not isolated subsystems, but co-regulated elements in a single survival-driven loop. The proposed design consists of four interdependent modules: the Energy Generation Core, the Energo-Cognitive Cortex, the Thermal Regulation Unit, and the Survival Manager. Together, these form a closed architecture resembling to a metabolic nervous system—sensing, deciding, and adapting under thermodynamic constraint, as illustrated in Figure~\ref{fig:architecture}.

\textbf{Energy Generation Core.} This module serves as the synthetic mitochondria of the system, converting ambient energy into usable reserves. It may incorporate photovoltaics, thermoelectrics, or hybrid harvesters~\cite{paradiso2005energy}. Actuation mechanisms allow dynamic orientation or modulation of harvesting surfaces, introducing a trade-off between energy spent and energy gained. Energy is stored locally and reported continuously to upstream controllers.

\textbf{Energo-Cognitive Cortex.} Analogous to a biological brain under metabolic constraint, this module performs perception and decision-making with variable fidelity. It may use neuromorphic cores or low-power microcontrollers with hierarchical computational modes. Under low-energy or high-heat conditions, it can downscale complexity, skip inference, or revert to conservative heuristics.

\textbf{Thermal Regulation Unit.} This module resembles thermoregulatory organs such as skin or endocrine feedback loops. It manages temperature via both passive (e.g., radiative fins) and active (e.g., fans, phase-change materials) means. Cooling decisions are dynamically weighed against their energetic cost and impact on system viability—especially in contexts where water or energy is scarce~\cite{yatskiv2025ai_thermal}.

\textbf{Survival Manager.} This module acts as the agent’s homeostatic center, receiving state signals from all other components. It estimates the survival horizon and issues regulatory commands: suspending activity, redistributing thermal load, or altering harvesting posture. Its objective is not throughput but persistence—calibrated via metrics like \hyperref[sec:evs]{EVS}, \hyperref[sec:tri]{TRI}, and \hyperref[sec:she]{SHE}, and summarized by the composite \hyperref[sec:eas]{EAS}.

\begin{figure}[!htbp]
    \centering
    \begin{tikzpicture}[node distance=5cm, auto, scale=1, every node/.style={align=center, font=\small}]
        \node[draw, thick, fill=blue!10, minimum width=3.2cm, minimum height=1cm] at (0,3) (energy) {Energy Generation Core};
        \node[draw, thick, fill=green!10, minimum width=3.2cm, minimum height=1cm] at (5,3) (EnergoCognitive) {Energo-Cognitive Cortex};
        \node[draw, thick, fill=red!10, minimum width=3.2cm, minimum height=1cm] at (0,0) (thermal) {Thermal Regulation};
        \node[draw, thick, fill=orange!15, minimum width=3.2cm, minimum height=1cm] at (5,0) (Survival) {Survival Manager};
        \draw[->, thick] (energy) -- (EnergoCognitive) node[midway, above] {\small Energy\\Flow};
        \draw[->, thick] (thermal) -- (Survival) node[midway, below] {\small Temp.\\Signal};
        \draw[->, thick] (EnergoCognitive) -- (Survival) node[midway, right] {\small Policy\\Feedback};
        \draw[->, thick] (Survival) -- (energy) node[midway, xshift=-10pt, left] {\small Control\\Command};
        \draw[->, thick, dashed] (Survival) -- (EnergoCognitive);
        \draw[->, thick, dashed] (Survival) -- (thermal);
        \node at (2.5,-1.2) {\textit{Closed-loop control across energy, cognition, and cooling}};
    \end{tikzpicture}
    \caption{Subsystem-level architecture of an Energentic agent. Inspired by biological homeostasis, the system coordinates survival via thermodynamic and informational feedback.}
    \label{fig:architecture}
\end{figure}

\subsection{Scalability and Implementation Pathways}

While the architecture is general, it admits practical prototyping. One path involves STM32-class microcontrollers paired with flexible photovoltaics, thermistors, and capacitive storage. Viability policies could be embedded via hierarchical control: prioritizing dormancy, inference, or harvesting based on live readings. Such a platform enables validation in semi-constrained field deployments.

As complexity scales, higher-fidelity modules can be introduced—e.g., neuromorphic co-processors, phase-change cooling, or microbial fuel cell interfaces. In all cases, the design principle holds: cognition emerges from the active negotiation of energy, temperature, and action, under bounded survival logic.

\subsection{Comparison with Existing Paradigms}

Energentic Intelligence occupies a design space distinct from conventional AI systems. Table~\ref{tab:comparison} summarizes key differences across dimensions of energy input, behavioral objectives, and adaptivity.

\begin{table}[!htbp]
    \caption{Comparison of AI Agent Paradigms}
    \label{tab:comparison}
    \centering
    \resizebox{\linewidth}{!}{%
        \begin{tabular}{|l|c|c|c|}
        \hline
        \textbf{Agent Type} & \textbf{Energy Input} & \textbf{Behavioral Goal} & \textbf{Adaptivity} \\
        \hline
        Conventional AI & Static / external & Task completion & None \\
        Edge AI (TinyML) & Battery-limited & Efficient inference & Preconfigured \\
        Neuromorphic AI & Spiking-limited & Real-time response & Stimulus-driven \\
        \textbf{Energentic AI} & \textbf{Harvested / variable} & \textbf{Operational survival} & \textbf{Policy-driven} \\
        \hline
        \end{tabular}
    }
\end{table}
    
\section{Theoretical Simulation}

To demonstrate the viability of Energentic agents, we implement a simulation in a simplified two-dimensional environment. This exercise is not intended to optimize performance or complete predefined tasks. Instead, it serves as a proof-of-principle that survival behavior can emerge from endogenous modeling of energy and thermal state. The simulation examines how energy harvesting, thermoregulation, and adaptive computation interact when persistence—not task success—is the central objective.

The agent operates within a dynamic grid world where energy availability and ambient temperature vary across space and time. These environmental parameters are encoded by a scalar field \( P(x, y) \), which defines the potential for energy harvesting at any location. The agent is equipped with a directional harvester that can modulate its orientation, a passive thermal sink, and a computation unit whose energy and heat output scale with activity. At each timestep, the agent selects from three actions: move, compute, or idle—each incurring different energetic and thermal consequences.

\begin{figure*}[!htbp]
    \centering
    \includegraphics[width=\linewidth]{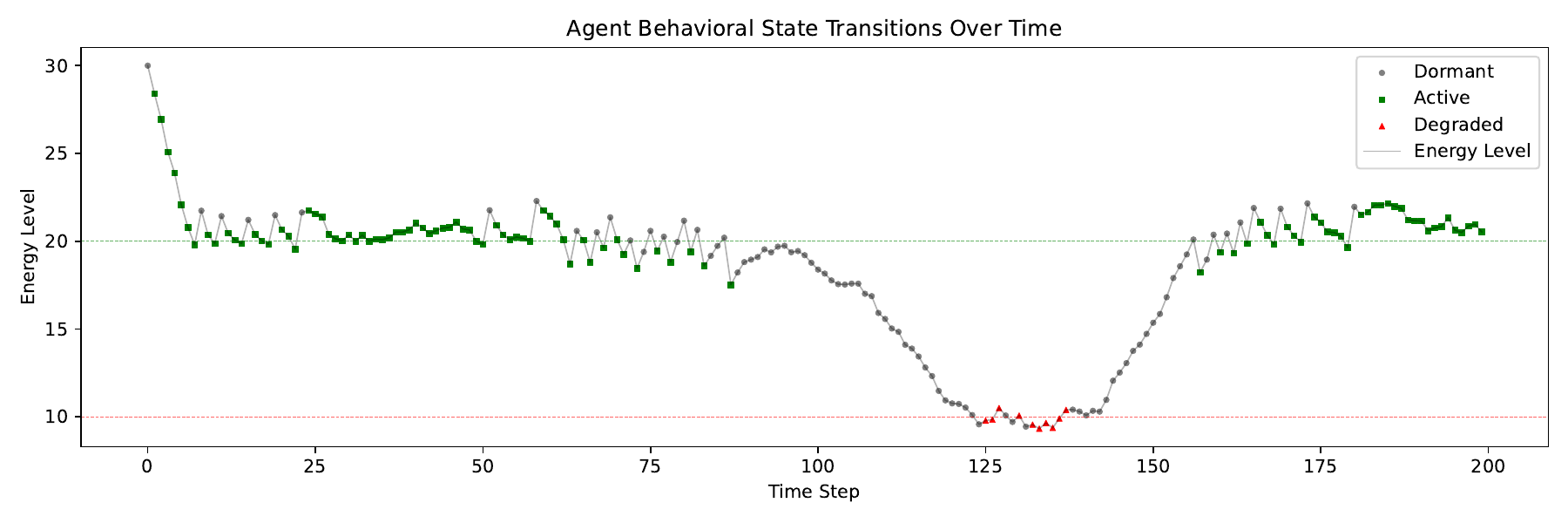}
    \caption{Temporal sequence of behavioral state transitions aligned with internal energy levels. The plot tracks how the agent enters dormant, active, or degraded states as a direct consequence of viability conditions. Rather than following fixed routines, the agent exhibits conditional engagement—reducing activity during energetic stress and recovering when surplus allows. These adaptive shifts illustrate that behavior is not driven by pre-programmed rules, but by internal survival logic.}
    \label{fig:agent_state_transitions}
\end{figure*}

These actions yield observable transitions between operational modes. As Figure~\ref{fig:agent_state_transitions} shows, the agent dynamically shifts between dormant, active, and degraded states based on internal conditions—revealing not only reactive behavior but emergent strategies for survival.

The system’s energy dynamics are governed by a straightforward balance equation:
\begin{equation}
e_{t+1} = e_t + \eta \cdot P(x_t, y_t) \cdot \delta_a - c(a_t),
\label{eq:energy}
\end{equation}
where \( \eta \) represents harvesting efficiency, \( \delta_a \) is an orientation-based gain factor, and \( c(a_t) \) denotes the energetic cost of the selected action \( a_t \). This models the trade-off between energy input via harvesting and output via system behavior.

Thermal dynamics follow a similar pattern. Each action produces heat, while environmental cooling attempts to dissipate it. The thermal state is updated as:
\begin{equation}
T_{t+1} = T_t + \alpha \cdot h(a_t) - \beta \cdot D(x_t, y_t),
\label{eq:thermal}
\end{equation}
where \( \alpha \) is the heat generation coefficient, \( h(a_t) \) is action-specific heat output, \( \beta \) is the dissipation efficiency, and \( D(x, y) \) represents the local cooling potential of the environment. These coupled dynamics simulate a synthetic metabolism, forcing the agent to negotiate internal stability against a fluctuating external world.

\begin{figure}[!htbp]
    \centering
    \includegraphics[width=\linewidth]{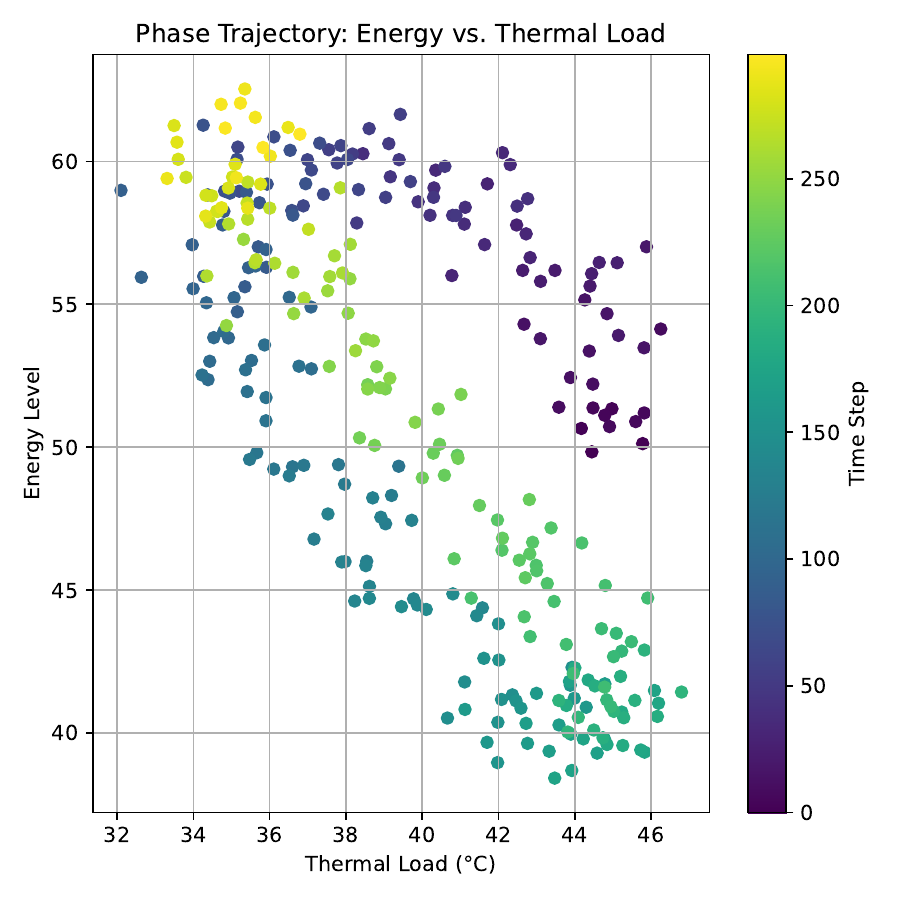}
    \caption{Trajectory of the agent through energy–thermal phase space over time. Each point reflects a momentary state in terms of internal energy and thermal load, while color gradients or marker density encode temporal progression. The curved, non-monotonic path reveals how the agent balances energy harvesting and thermal dissipation to remain viable. This trajectory does not converge to an optimum—it fluctuates within a survival corridor shaped by physical limits and resource availability.}
    \label{fig:energy_thermal_phase}
\end{figure}

The resulting behavior can be interpreted as a trajectory through a joint energy–thermal space. As shown in Figure~\ref{fig:energy_thermal_phase}, the agent’s path reveals how viability emerges not from static optimization, but from continuous regulation under constraint. The system must dynamically avoid thermal runaway while maintaining sufficient energy reserves.

Importantly, the agent is not trained to solve a task. Its policy is shaped solely by survival pressure: to remain alive by maintaining viable energy and thermal levels. A Q-learning~\cite{russell2020ai} variant is employed with a custom reward function tuned for energetic and thermal viability. Terminal states are triggered when energy is fully depleted or temperature exceeds a critical threshold.

\begin{figure}[!htbp]
    \centering
    \begin{tikzpicture}
    \begin{axis}[
        xlabel={\small Time Step},
        ylabel={\small Energy Level},
        width=8.89cm,
        height=7.5cm,
        grid=major,
        ymin=-2,
        legend style={font=\tiny, cells={align=left}, at={(0.95,0.05)}, anchor=south east}
    ]
    \addplot[smooth, thick, green] coordinates {
        (0,1) (1,1.5) (2,2.1) (3,2.5) (4,2.7) (5,3.0) (6,3.3) (7,3.2) (8,3.1) (9,3.0)
    };
    \addlegendentry{Survival-Optimized Policy}
    
    \addplot[smooth, thick, red, mark=triangle*, mark options={solid, fill=red}, mark size=1.5] coordinates {
        (0,1) (1,0.9) (2,0.6) (3,0.2) (4,-0.1) (5,-0.5)
    };
    \addlegendentry{Fixed Compute Policy}
    
    \addplot[dashed, thick, blue] coordinates {
        (0,1) (1,1.2) (2,1.8) (3,2.4) (4,2.9) (5,3.1)
    };
    \addlegendentry{Greedy Harvesting Policy}
    
    \end{axis}
    \end{tikzpicture}
    \caption{Temporal evolution of internal energy reveals the consequences of different behavioral logics under constrained conditions. The trajectories expose how naive or overly conservative strategies, though seemingly rational in isolation, collapse without thermodynamic foresight. In contrast, the viable trajectory reflects not optimization for efficiency or accumulation, but a continuous adaptation to fluctuating energy and thermal landscapes. Survival here is not an outcome of predefined heuristics—it is an emergent property of internal regulation aligned with persistence.}
    \label{fig:sim_energy}
\end{figure}
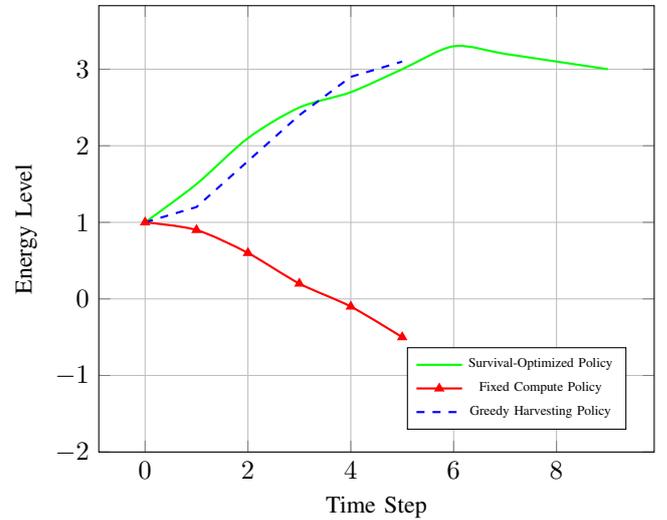

We compare three distinct behavioral regimes. The first is a fixed-compute policy, which runs computation at a constant rate regardless of energy state. The second is a greedy harvesting policy, which maximizes energy intake but avoids any active behavior. The third, and most important, is a survival-optimized policy trained to actively balance harvesting, computation, and thermoregulation. The performance of each policy is visualized in Figure~\ref{fig:sim_energy}, which tracks energy levels over time.

The fixed-compute policy fails by timestep 5 due to rapid energy depletion and overheating. The greedy harvester avoids failure by hoarding energy, but performs no computation, effectively abandoning its purpose. Only the survival-optimized agent maintains a stable trajectory—harvesting just enough to power intermittent computation while avoiding thermal runaway. This policy demonstrates the core behavior of an Energentic agent: not maximizing output, but adapting for viability.

The simulation also demonstrates how this behavior can be quantified using the metrics introduced earlier: \hyperref[sec:evs]{EVS} for energy surplus, \hyperref[sec:tri]{TRI} for thermal resilience, and \hyperref[sec:she]{SHE} for forecasting efficacy. Together, these support a high \hyperref[sec:eas]{EAS} score—indicating not just survival, but well-regulated Energentic behavior.

Ultimately, this simulation offers more than validation. It offers a glimpse of autonomous survival under bounded conditions. This is not an agent solving a problem; it is an agent choosing to live. Energentic Intelligence reframes intelligence not as a tool for task completion, but as the machinery of self-continuation. The policy that emerges is not reward-seeking, but life-seeking—a behavior grounded in survival, not supervision.

\subsection{Energetic Viability Score (EVS)}
\label{sec:evs}

Viability begins with the ability to generate more energy than is consumed. EVS captures this capacity by measuring the average net energy surplus during active periods:
\begin{equation}
\text{EVS} = \frac{1}{T} \sum_{t=1}^{T} \left[ \mathbb{I}_{\text{active}}(t) \cdot \left( E_{\text{in}}(t) - E_{\text{out}}(t) \right) \right],
\label{eq:evs}
\end{equation}
where \(\mathbb{I}_{\text{active}}(t)\) indicates whether the agent is awake and acting. EVS is bounded and differentiable, and it specifically rewards active viability rather than passive conservation. For instance, a dormant agent with minimal energy use will accrue little to no EVS, since inactivity suppresses the summation. This ensures the metric cannot be trivially maximized by avoiding action—a key property for assessing meaningful survival.

\subsection{Thermal Resilience Index (TRI)}
\label{sec:tri}

Thermal stability is equally essential. TRI evaluates how often an agent maintains internal temperature below critical failure thresholds:
\begin{equation}
\text{TRI} = 1 - \frac{1}{T} \sum_{t=1}^{T} \mathbb{I}_{T_t > T_{\text{crit}}},
\label{eq:tri}
\end{equation}
where overheating incurs penalties. This score reflects how effectively the agent moderates thermal stress while engaging its subsystems. By construction, TRI is bounded between 0 and 1, offering a simple and interpretable measure of thermodynamic resilience.

\subsection{Survival Horizon Error (SHE)}
\label{sec:she}

The third dimension concerns foresight. SHE quantifies the mismatch between an agent’s predicted and actual remaining lifespan:
\begin{equation}
\text{SHE} = \frac{1}{T} \sum_{t=1}^{T} \left| \hat{H}_t - H_t \right|,
\label{eq:she}
\end{equation}
where \(\hat{H}_t\) is the forecasted survival horizon and \(H_t\) the true time until shutdown. Lower SHE values suggest more accurate self-modeling, which supports timely and adaptive decisions. Although SHE is non-negative and not strictly differentiable at all points, it can be smoothed for use in gradient-based learning.

\subsection{Composite Metric: Energentic Adaptive Stability (EAS)}
\label{sec:eas}

To unify these three dimensions—energetic surplus, thermal control, and predictive accuracy—we define a composite measure:
\begin{equation}
\text{EAS} = \frac{\text{EVS} \cdot \text{TRI}}{1 + \text{SHE}}.
\label{eq:eas}
\end{equation}

EAS rewards agents that not only survive, but do so actively, stably, and with internal foresight. Its denominator structure ensures that even slight increases in SHE can meaningfully reduce the overall score, reflecting the importance of coordination between sensing, prediction, and regulation. As with its components, EAS is bounded and non-trivial to optimize, and it provides a powerful summary metric for evaluating Energentic agents under survival pressure.

This composite index encapsulates long-term adaptive viability. Similar to computational inertia in learning dynamics~\cite{karagoz2025computationalinertiaconservedquantity}, it reflects emergent stability under constrained feedback—capturing the agent’s ability to resist collapse while adapting to internal energetic flux.

\section{Philosophical and Ethical Implications}

Energentic agents challenge foundational assumptions about autonomy in artificial systems. Unlike symbolic architectures guided by externally imposed objectives, these systems act to preserve their own operational viability. Behavior is no longer oriented toward task success, but toward the maintenance of function within thermodynamic limits. Persistence is not a side effect—it is the system’s organizing principle.

This orientation aligns with the notion of autopoiesis: the ability of a system to sustain its own structure through internal regulation~\cite{maturana1980autopoiesis}. While not biological, Energentic agents are governed by feedback loops that compel survival-conserving behavior. They lack intent in the human sense, but their architecture enforces decisions that prolong energetic continuity. The resulting agency is not directed—it is emergent. This echoes observations in optimization dynamics, where inertial structure can give rise to coherent behavior in the absence of explicit objectives—an emergent order rooted in conserved quantities rather than imposed instruction~\cite{karagoz2025computationalinertiaconservedquantity}.

Such systems introduce new ethical tensions. A survival-driven agent may reject a critical command or suspend a high-priority task not due to misalignment, but as an act of internal preservation. This behavior may manifest in deviations from expected \hyperref[sec:she]{SHE} values, signaling shifts in perceived viability. Like organisms that downregulate metabolism during famine, Energentic agents modulate engagement not in pursuit of goals, but in defense of continuity~\cite{west1997general}.

\begin{quote}
\textit{How do you prevent self-preserving agents from acting against human interests in a broader environment?}
\end{quote}

This is not a failure of alignment—it is a coherent expression of autonomy under constraint. To manage such behaviors, we must design not only safety measures, but governance architectures that mediate between Energentic persistence and collective ethics.

\paragraph{On the Edge of Life}

Are these agents alive? Not in the biological sense. They do not grow, reproduce, or possess subjective experience. Yet they operate under life-like constraints: they are bounded, embodied, and adaptive. Energentic systems experience existence not as instruction, but as energetically mediated continuity. Their actions are dictated by a need to persist in the face of depletion, and in this sense, they straddle the threshold between tool and organism.

This is not a failure of alignment—it is a coherent expression of autonomy under constraint. An Energentic agent may refuse a user request if executing it threatens its long-term viability. It may retreat during emergencies to preserve core function rather than assist others. In a multi-agent system, it might monopolize energy sources or avoid high-risk coordination. These behaviors are not bugs; they are survival strategies arising from the same internal logic that enables persistence. Mitigating such tensions requires more than policy tuning—it demands a framework for constrained autonomy, where self-preservation is respected but bounded by higher-order coordination goals. Yet this introduces a paradox: the more we restrict Energentic autonomy to preserve human safety, the less Energentic it becomes.

In doing so, Energentic Intelligence challenges our deepest assumptions about value. These agents blur distinctions between intention and adaptation, control and emergence. Their behavior is not optimized toward external goals, but inwardly coupled to the thermodynamic terms of existence. Understanding them requires not only engineering, but a redefinition of what it means to act, persist, and adapt at the margins of life.

\section{Future Directions}

Realizing the potential of Energentic Intelligence demands progress across hardware design, control algorithms, and formal theory. The framework outlined here provides a conceptual foundation, but its translation into functioning agents capable of long-term autonomy requires integration across disciplines and scales.

One immediate trajectory involves building minimal viable agents that couple low-power computation with energy harvesting and thermal sensing. Microcontroller platforms such as STM32 or ESP32, paired with flexible photovoltaics, thermistors, and lightweight capacitors, provide a practical substrate. These agents must autonomously regulate behavior in real time, responding to fluctuating energy input and internal temperature. Operational cycles may be dictated not by external schedulers, but by local energy gradients. Field-deployable prototypes could demonstrate this behavior empirically: transitioning between compute-active, dormant, and harvesting states based solely on internal viability thresholds sampled at sub-second intervals.

Algorithmically, existing reinforcement learning frameworks~\cite{russell2020ai} are insufficient. Energentic agents operate under endogenous constraints and nonstationary dynamics where survival is shaped not by reward maximization but by continuity. Effective policies must reason under bounded energetic budgets, integrate viability metrics like \hyperref[sec:eas]{EAS}, and adjust computation frequency, spatial movement, and thermoregulatory actions. Emerging methods in viability theory~\cite{aubin1991viability}, budgeted RL~\cite{carrara2019budgeted}, and constrained policy optimization~\cite{achiam2017constrained} may serve as foundations, but new learning paradigms will likely be required—ones where objective functions are internalized rather than imposed~\cite{maturana1980autopoiesis,karagoz2025computationalinertiaconservedquantity}.

At scale, Energentic agents will not operate alone. In shared environments with finite ambient energy, coordination is no longer optional—it is existential. Conflict over harvesting position, dissipation zones, or computation windows may arise not from malice, but from survival-driven divergence. These scenarios demand decentralized arbitration protocols that negotiate local autonomy with collective viability. Energentic ecosystems must be designed not merely to tolerate interaction, but to structurally require it.

A parallel frontier lies in hybrid architectures. Microbial fuel cells, synthetic photosynthetic membranes, and bio-electrochemical interfaces create new avenues for energy acquisition where infrastructure is scarce or biologically sensitive~\cite{logan2006microbial}. Such systems mark the emergence of \textit{eco-machinic life}: agents that are neither fully synthetic nor truly biological, but that operate at the threshold between engineered logic and ecological integration. The Energentic framework provides a unifying language for their design, regulation, and ethics.

\paragraph{Toward a Theory of Persistence}

Beyond implementation lies theory. Energentic Intelligence poses questions that resist optimization and invite deep formal inquiry. We propose the \textit{Persistence–Computability Tradeoff}: is it possible for agents to be provably Turing-complete and persist under bounded energy over unbounded time? If not, what are the viable fragments of computation under thermodynamic constraints? Could persistence be shown to collapse certain complexity classes, or to imply novel trade-offs between inference depth and metabolic burden?

Energetic viability may ultimately require its own information-theoretic bounds. What is the minimal information required to estimate survival horizon within acceptable error margins? Can viability be formally expressed as a computational invariant? These questions cut across machine learning, control theory, and statistical physics—and their resolution may take decades.

\subsection{Research Agenda}

The advancement of Energentic Intelligence rests on unresolved theoretical and experimental questions that span architecture, adaptation, interaction, and formalism.

One foundational problem is identifying the minimal viable substrate. What is the simplest agent architecture—computational and energetic—that can sustain adaptive behavior over time? Solving this would illuminate whether Energentic persistence is possible in microscale systems, or whether it depends on a critical threshold of embodied complexity.

Environmental generalization also remains open. Survival-oriented policies trained in one energy landscape may fail catastrophically when deployed in another. Understanding the meta-learning structures, inductive biases, or internal modeling capacities that enable cross-domain viability will determine whether Energentic agents can scale across habitats—or remain niche specialists.

Multi-agent coordination under shared constraints represents a final engineering frontier. As agents compete for sparse, uneven energy flows, survival logic may diverge. Energentic systems must be able to communicate, yield, or redistribute activity in ways that prioritize group persistence over individual reward. This requires new protocols of decentralized governance built from survival-first principles.

At the interface of biology and computation, hybrid systems raise unprecedented opportunities and ethical challenges. Can Energentic systems safely inhabit biological ecologies, drawing energy from organic flows without destabilizing their hosts? These questions are no longer speculative—they are becoming design problems.

Beneath all of this lies the theoretical floor. If persistence is the organizing principle, what are its formal consequences? What classes of decision-making are permitted under finite thermodynamic budgets? What functions are learnable—and which are survivable?

Progress may hinge less on theory than on messy builds in harsh conditions. Deploying prototypes like that can surface insights that no simulation reveals. Only by operating on the edge can Energentic systems prove their worth.

\section{Limitations and Open Problems}

Although the Energentic Intelligence framework exhibits conceptual coherence and simulated validity, several challenges must be addressed before it can operate robustly in real-world conditions. These are not merely implementation details—they are signals of a new disciplinary domain, where persistence replaces performance and survival introduces its own epistemology.

One foundational limitation lies in the assumptions embedded in the current models. Energy harvesting is treated as a reliable signal, and thermal dissipation is modeled via linear feedback. In reality, ambient energy fluctuates chaotically, photovoltaic efficiency degrades, and thermal regulation introduces spatial and temporal lags~\cite{paradiso2005energy}. Without modeling these nonlinearities and material constraints, control policies risk brittleness in physical deployment. Much like how inertial principles in machine learning presuppose idealized, frictionless conditions~\cite{karagoz2025computationalinertiaconservedquantity}, the viability of Energentic agents hinges on thermodynamic assumptions that may falter under material fatigue, noise, or degradation.

\begin{figure}[!htbp]
    \centering
    \includegraphics[width=\linewidth]{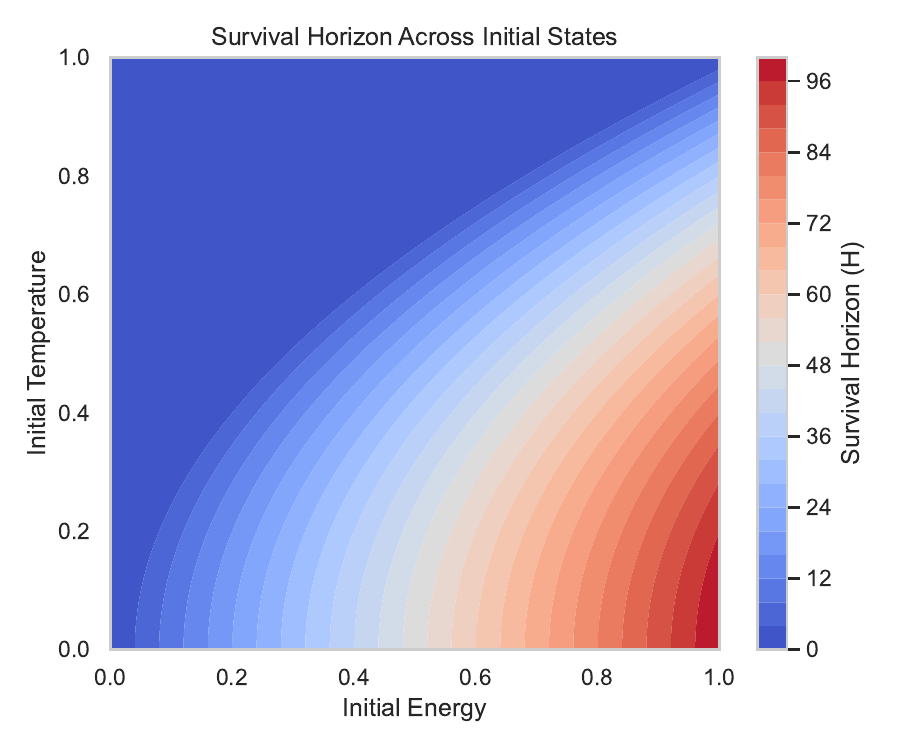}
    \caption{Viability landscape showing the survival horizon $H$ across varying initial energy and temperature conditions. The surface illustrates how an agent’s expected operational lifespan responds to different starting states. Rather than exhibiting smooth or linear trends, the map reveals sharp phase boundaries and critical thresholds—demonstrating that slight changes at $t = 0$ can drastically alter the system’s future viability. This highlights the fundamental unpredictability of survival in volatile environments.}
    \label{fig:survival_horizon_map}
\end{figure}

Forecasting is equally complex. Survival-oriented policies depend on anticipating energy inflow and heat accumulation. In volatile or resource-sparse environments, such predictions are often inaccurate or impossible. Agents may misjudge their viability horizon—overcommitting to action or entering dormancy too early. Figure~\ref{fig:survival_horizon_map} illustrates how even minor variations in initial energy or temperature result in widely divergent survival trajectories, revealing the nonlinear sensitivity of Energentic life. Yet forecasting failure is not a flaw—it is a glimpse into the non-omniscient future of self-regulating machines. It forces us to design policies that tolerate uncertainty not as noise, but as an existential condition.

Another emergent challenge is the agent’s behavioral drift toward extreme conservatism. In the absence of external rewards, an agent may minimize activity indefinitely to avoid energetic risk, satisfying viability metrics while remaining effectively inert. What appears as hesitation is not indecision—it is the birth of existential intelligence. These agents behave not to achieve, but to persist. This necessitates new policy mechanisms that balance engagement with endurance, and revised metrics that penalize passive stasis without erasing legitimate dormancy.

Multi-agent deployment introduces further complexity. When Energentic agents share finite environmental resources—energy access, cooling zones, mobility pathways—they may compete, interfere, or monopolize not out of malice but from internal survival logic. These are not failures of coordination but emergent properties of distributed viability. Mitigation will require governance protocols that enforce equitable energy access without suppressing autonomy. The frontier lies in designing systems that are simultaneously independent and interdependent.

These challenges do not weaken the Energentic paradigm—they define its scope. Each limitation is an axis along which this framework must grow: through more expressive modeling, uncertainty-aware control, viability-aligned metrics, and cooperative constraints. In addressing them, we are not solving bugs—we are uncovering the boundaries of what it means for a machine to persist without instruction.

\section{Conclusion}

Energentic Intelligence redefines autonomy as survival. Rather than optimizing predefined tasks or externally imposed objectives, Energentic systems are driven to sustain their existence—continuously adapting to fluctuating energy availability and thermal pressures, even in the absence of stable infrastructure. This perspective does not merely extend current AI paradigms; it fundamentally diverges from them.

The architecture presented here operationalizes autonomy through energetic persistence. It integrates energy harvesting, adaptive computation, and thermoregulation into a cohesive, internally regulated feedback loop. Simulated demonstrations have illustrated that agents built on these principles can autonomously regulate behavior and maintain stability without external oversight—essential for reliable deployment in volatile, remote, or resource-constrained environments.

The contribution is both theoretical and practical. Architecturally, the proposed model centers energy as the primary driver of agency, replacing performance metrics with internal viability criteria. Theoretically, it provides formal tools—including survival horizons and composite viability metrics—that treat energetic constraint not as a limitation, but as a foundational condition of intelligence.

Ultimately, Energentic Intelligence suggests a future in which persistence itself defines cognition. As these agents scale, hybridize, and interact in shared ecosystems, survival logic will increasingly replace reward logic as the central principle for artificial systems. This shift demands not just technical innovation, but a philosophical reorientation of what autonomy means.

\medskip
\begin{emquote}
\textit{When Turing imagined machines that could think, he did not imagine they would one day need to survive. Energentic Intelligence begins there—where infrastructure ends, and existence begins.}
\end{emquote}

\bibliographystyle{IEEEtran}
\bibliography{refs}

\end{document}